\documentclass[11pt, a4paper,  notitlepage]{extarticle}

\usepackage[left=3.5cm, right=2.5cm, top=3cm, bottom=3cm, bindingoffset=0cm]{geometry}
\usepackage{graphicx}
\usepackage{mathtools}
\usepackage[affil-it]{authblk}
\usepackage{url}
\usepackage{amsmath}
\usepackage{amssymb}
\usepackage{multicol}
\usepackage{float}
\usepackage[section]{placeins}
\usepackage{hyperref}
\usepackage[title]{appendix}

\begin{document}
\sloppy
\title{Analysis of Disinformation and Fake News Detection Using Fine-Tuned Large Language Model}
\author{Bohdan M.  Pavlyshenko \\  \small{Ivan Franko National University of Lviv,  Ukraine \\  b.pavlyshenko@gmail.com,  www.linkedin.com/in/bpavlyshenko/ }}
\maketitle

\begin{abstract}
The paper considers the possibility of fine-tuning Llama 2 large language model (LLM) for the disinformation analysis and fake news detection.  For fine-tuning, the PEFT/LoRA based approach was used. In the study, 
the model was fine-tuned for the following tasks: analysing a text on revealing disinformation and  propaganda narratives, fact checking, fake news detection,  manipulation analytics, extracting named entities with their sentiments.
The obtained results show that the fine-tuned Llama 2 model can perform a deep analysis of texts and reveal complex styles and narratives.  Extracted sentiments for named entities can be considered as predictive features in supervised machine learning models. 

Keywords: fake news detection,  Twitter, news trends, frequent itemsets, transformers, deep learning, users' communities, Large Language Model, Llama 2, LLM fine-tuning, PEFT, LoRA.
\end{abstract}

\tableofcontents

\section{Introduction}
Disinformation and fake news are amongst the top main problems nowadays. 
There are different approaches to producing disinformation, e.g. 
using not real  or not essential facts  which lead to specified conclusions;
 using offensive, emotional style with incorrect and biased accents on facts and other approaches.
News has an essential impact in many areas of society, politics and business. That is why one  can see a lot of attempts to produce manipulative and fake  news to get a specified response  in the society.  One of  horrible world events  is Russian invasion of Ukraine  on February 24, 2022.  It causes a large informational news flow on social networks, including producing manipulative and fake news to shape a specified explanation and justification of invasion. 

 Large Language Models (LLM)  due to their transformer structure with attention mechanism can help analyse complex texts and reveal text styles. 
LLM such as ChatGPT  show high efficiency in the analysis of complex texts. 
Nowadays, we can observe the emerging of many new smaller open source LLMs, e.g. Llama, Falcon, GPT4All, GPT-J, etc. 
Open source LLMs can be fine-tuned for specific custom problems and deployed on custom servers, e.g. in cloud computing services such as AWS, GCP. 
 LLMs have some new features as compared to conventional language models based on transformers. One of them is zero-shot and few-shot learning, which 
 consists in good performance of the model when we show it only few training examples or even no examples at all, but only the instructions describing what should be done. Another important feature is the reasoning when a model can generate new patterns and conclusions which are based on an input prompt and facts known by the model and which were not included into it directly during a training process. So, the model can generate analytical texts with unexpected but useful chains of thoughts.  One of the approaches of using LLMs is based on retrieval augmented generation (RAG), which uses the results from other services, e.g. relational database, semantic search, graph database in the input prompt for LLM. In this case,   the response  can be treated as the combination of  external results and LLM knowledge. 
In \cite{pavlyshenko2022methods},  different approaches for the analysis of news trends on Twitter have been considered.  
The obtained results show that an effective system for detecting fake and manipulative news can be developed using combined neural network which consists of three concatenated subnetworks. Discussions on social networks about companies' behavior have some impact on their business and their stock prices on the stock market. To analyze such an impact and make risk assessment, Bayesian regression can be used. Using the theory of frequent itemsets and association rules along with thematic fields of keywords, makes it possible to reveal the semantic structure for entities in news messages. 
LLMs are being effectively used in analysing  financial data and news. 
In~\cite{pavlyshenko2023financial},  we use the  fine-tuned Llama 2 LLM model for financial news analytics.
We study the possibility to fine-tune Llama 2 Large Language Model (LLM) for the multitask analysis of financial news. For fine-tuning, the PEFT/LoRA based approach was used. 
The obtained results show that the fine-tuned Llama 2 model can perform a multitask financial news analysis with a specified structure of response, part of response can be a structured text and another part of data can have JSON format  for further processing. 

In this study, we are going to consider and test the fine-tuned Llama 2 LLM model~\cite{touvron2023llama}  on news datasets and propaganda narratives, 
highlighting the main points of a text, summarizing this text and extracting named entities with appropriate sentiments. 
The main goal of this study is to consider the possibility of using a fine-tuned Large Language Model (LLM) for detecting and analysing disinformation, fake news, propaganda narratives and manipulations in news messages.

\section{Methods of Fake News Detection on Twitter}
Let us consider some of our previous results on the methods of informational trends analytics and fake news detection in tweets.
In the paper~\cite{pavlyshenko2022methods}, we consider different approaches to the analysis of news trends and detecting fake news on Twitter. For the analysis and case study,  informational trends on Twitter caused by Russian invasion of Ukraine in 2022  have been studied. One of the goals claimed by Russia  was the  'denazification' of Ukraine. One of the allegations of Russian propaganda was  that Ukraine was developing the biological weapon in special laboratories.  A deep learning approach for fake news detection has been analyzed. The use of the theory of frequent  itemsets and association rules,   graph theory for news trends analytics has been considered.  

Tweets, the messages of Twitter microblogs,  have high density of semantically important keywords.  Different studies on fake news detection in Twitter are considered in the papers~\cite{grinberg2019fake,ajao2018fake,helmstetter2018weakly}. 
In~\cite{pavlyshenko2019forecasting,pavlyshenko2019cantwitter, pavlyshenko2021forming}, we study different approaches for the analysis of messages on Twitter, as well as the use of  tweet features for forecasting different kinds of events. 
The work~\cite{pavlyshenko2022forming} considers a number of approaches for forming different predictive features of tweet data sets and using them in the predictive analysis for the decision-making support. 
 
As fake news,  we will consider the news information which is not true as well as the information which  can contain  real facts, but with incorrectly specified accents, and the focuses that lead to distorted conclusion and incorrect understanding of underlying processes.  For our analysis, we considered informational trends caused by Russian invasion of Ukraine in 2022. 
In the study, we also consider the possible impact of informational trends on different companies working in Russia during this conflict. 

\begin{figure}
\center
 \includegraphics[width=0.85\linewidth]{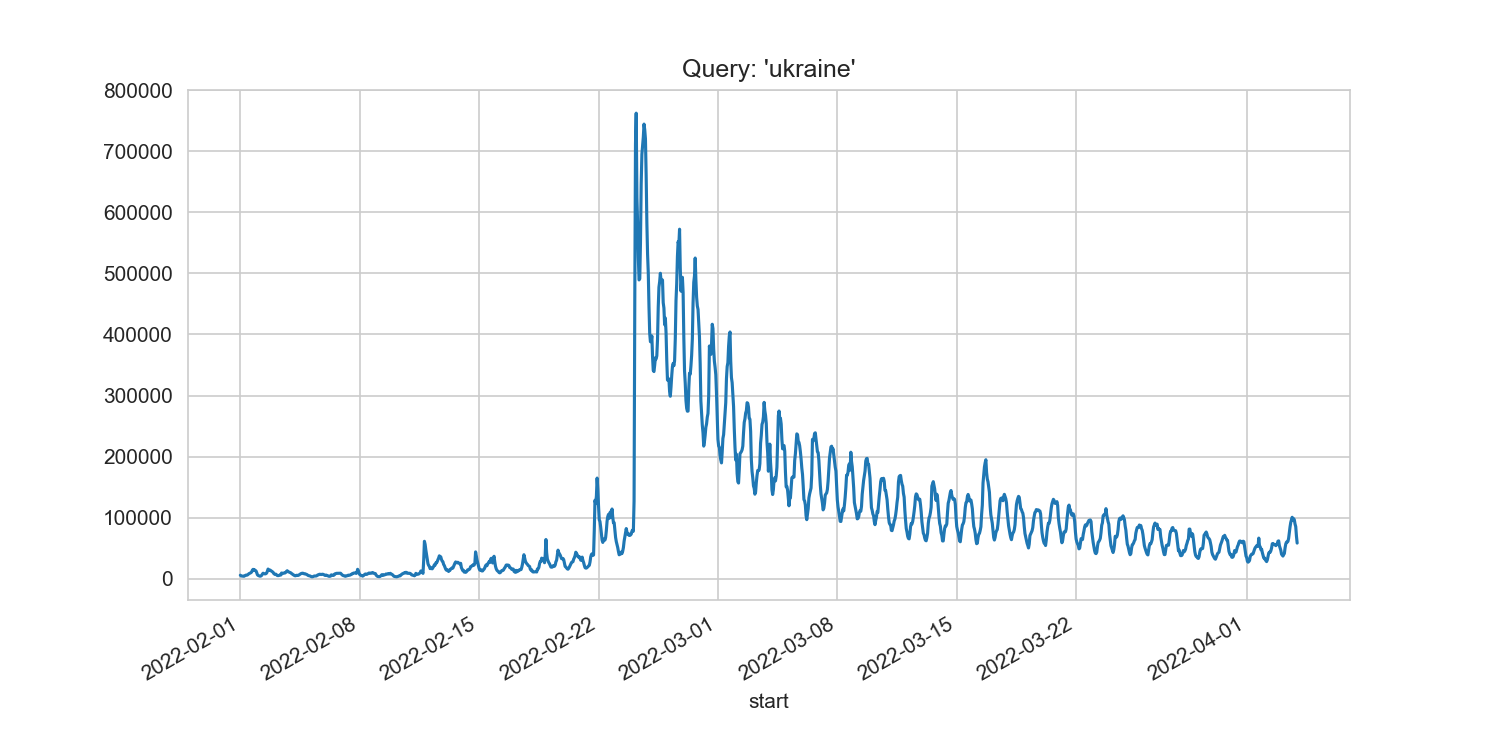}
 \caption{Time series of tweets for the query 'ukraine'.}
 \label{ts1}
 \end{figure}

\begin{figure}
\center
 \includegraphics[width=0.85\linewidth]{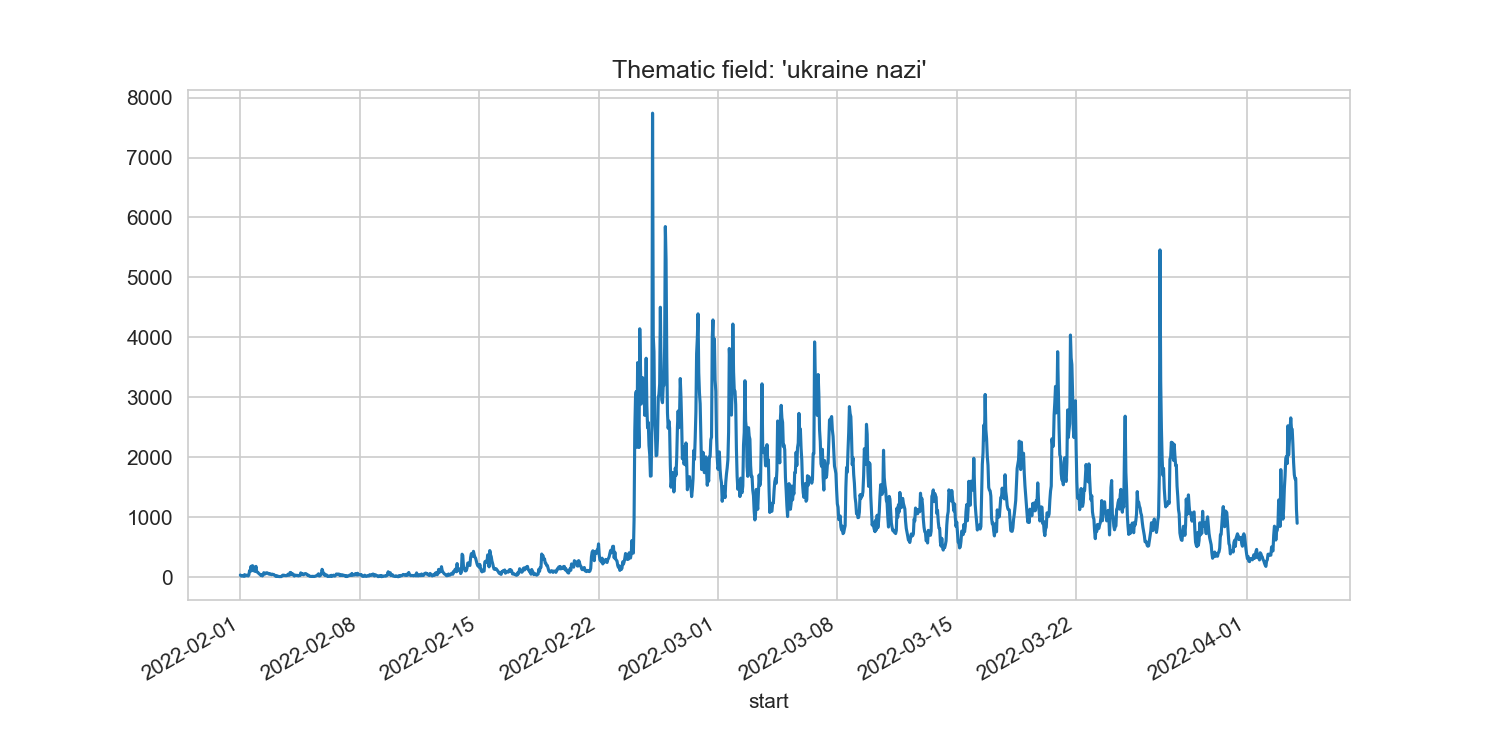}
 \caption{Time series of tweets for the thematic field 'ukraine nazi'.}
 \label{ts2}
 \end{figure}

\begin{figure}
\center
 \includegraphics[width=0.85\linewidth]{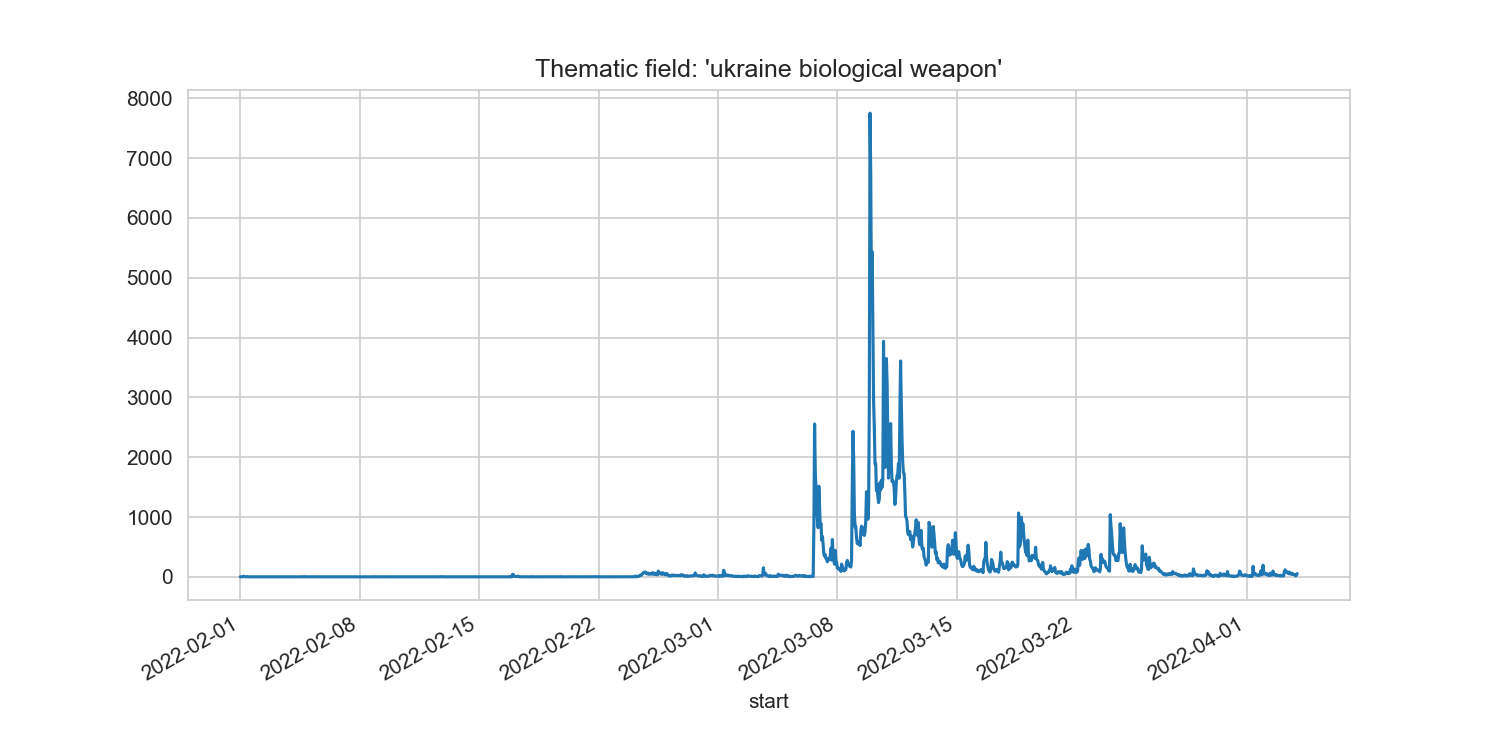}
 \caption{Time series of tweets  for the thematic field 'ukraine biological weapon'. }
 \label{ts4}
 \end{figure}

 Figures~\ref{ts1}-\ref{ts4} show the time series for tweet counts for different queries.  
As the results show,  for the 'ukraine nazi' thematic field, the discussion  of underlying problems   rose dramatically after February 24,  the date of Russian invasion of Ukraine. 
The amount of tweets related to this theme  before that date was at the minimum level. That itself leads to the conclusion that the problem with nazi in Ukraine was just a formal reason to justify the invasion.  Another  claim of Russian propaganda was about biological weapons that were allegedly being developed in Ukrainian laboratories (Figure~\ref{ts4}).
For instance,  it was claimed that a special virus was being developed and it was meant to be distributed through bats and migratory birds.  
\FloatBarrier

 
\begin{figure}
\center
 \includegraphics[width=0.95\linewidth]{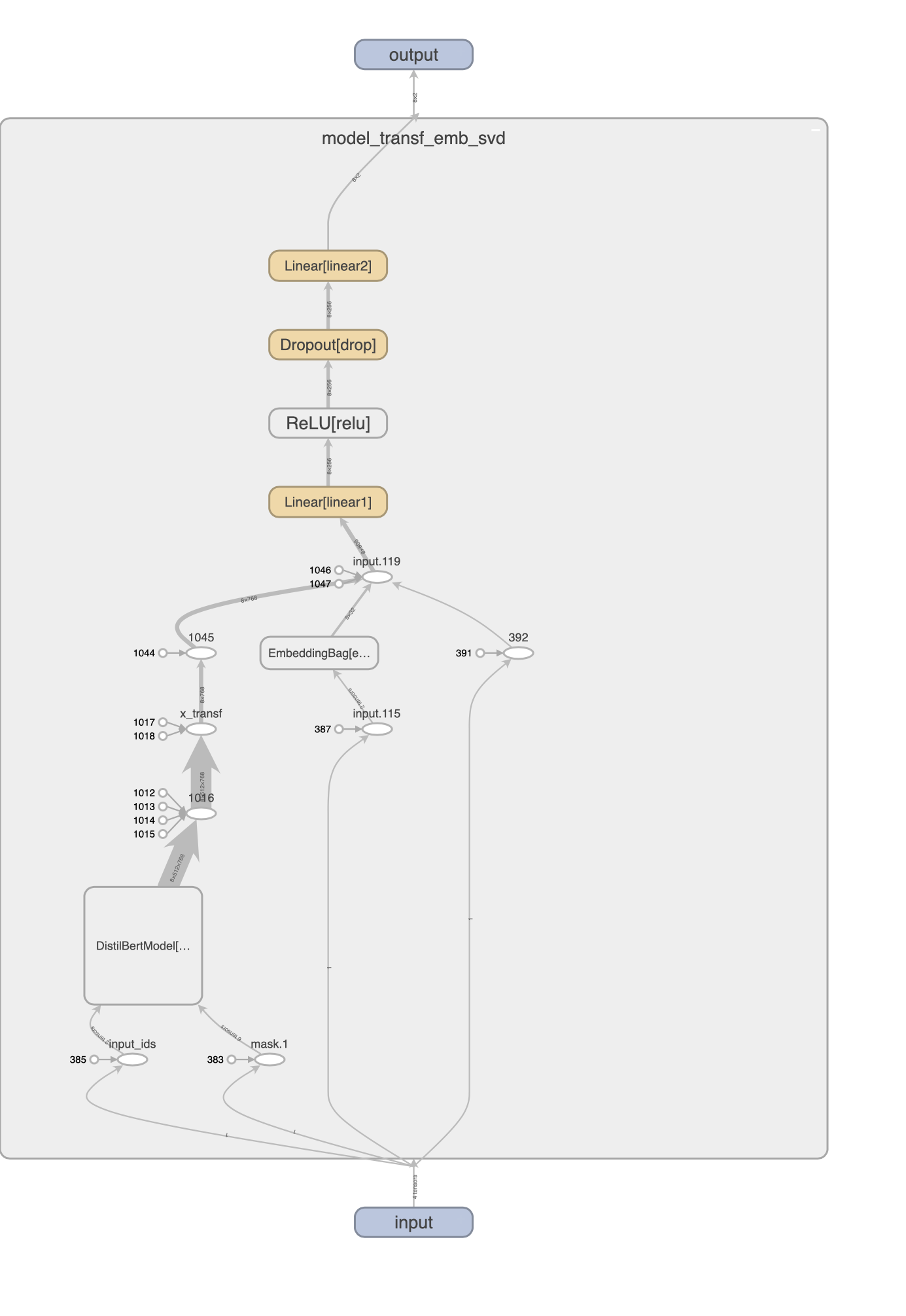}
 \caption{Deep learning model structure.}
 \label{fnmodel}
 \end{figure}

Let us consider the deep learning approach we used for fake news detection in tweets. 
Fake news can be detected and revealed  by analyzing  facts and comparing them with reality and other news sources. But for manipulative news, it is typical  to  amplify them artificially in different ways,  e.g. by retweeting manipulative tweets  many times using different users' accounts. Some accounts can be  bots which were artificially created, others can belong to real users. It makes it possible to detect fake news using an approach which analyzes the patterns of users' behavior.  Also, fake news have specific patterns in the text of messages. Both users' behavior and text patterns can be captured by deep machine learning algorithms. As the features for a predictive model, we used tweet texts and  the  list of users' usernames  who retweeted those tweets. The ML model consists of several concatenated neural subnetworks: subnetwork with DistilBERT transformer which  ingests tokens of tweet texts,  subnetwork with the embedding layer with averaging  which ingests the mixture of encoded words of tweet texts and lists of usernames of retweeters, 
as well as a subnetwork for the components of truncated singular value decomposition of TF-IDF matrix for the list of usernames of retweeters.  Figure~\ref{fnmodel} shows the structure of the deep learning model for fake and manipulative news detection.  For our case study,  the loaded tweets with the thematic fields  'ukraine nazi' and 'ukraine biological weapon' were used. For the model training and evaluation, the  tweet datasets with a specified list of users who retweeted those tweets were  created. For the analysis, only the tweets with a specified threshold for retweet counts were included.  The dataset was labeled using an appropriate tweet id, usernames, hashtags of tweets which can be treated as fake or manipulative. On testing this model,  f1 score  on the validation dataset was 0.95~\cite{pavlyshenko2022methods} . 

\begin{figure}
\center
 \includegraphics[width=0.95\linewidth]{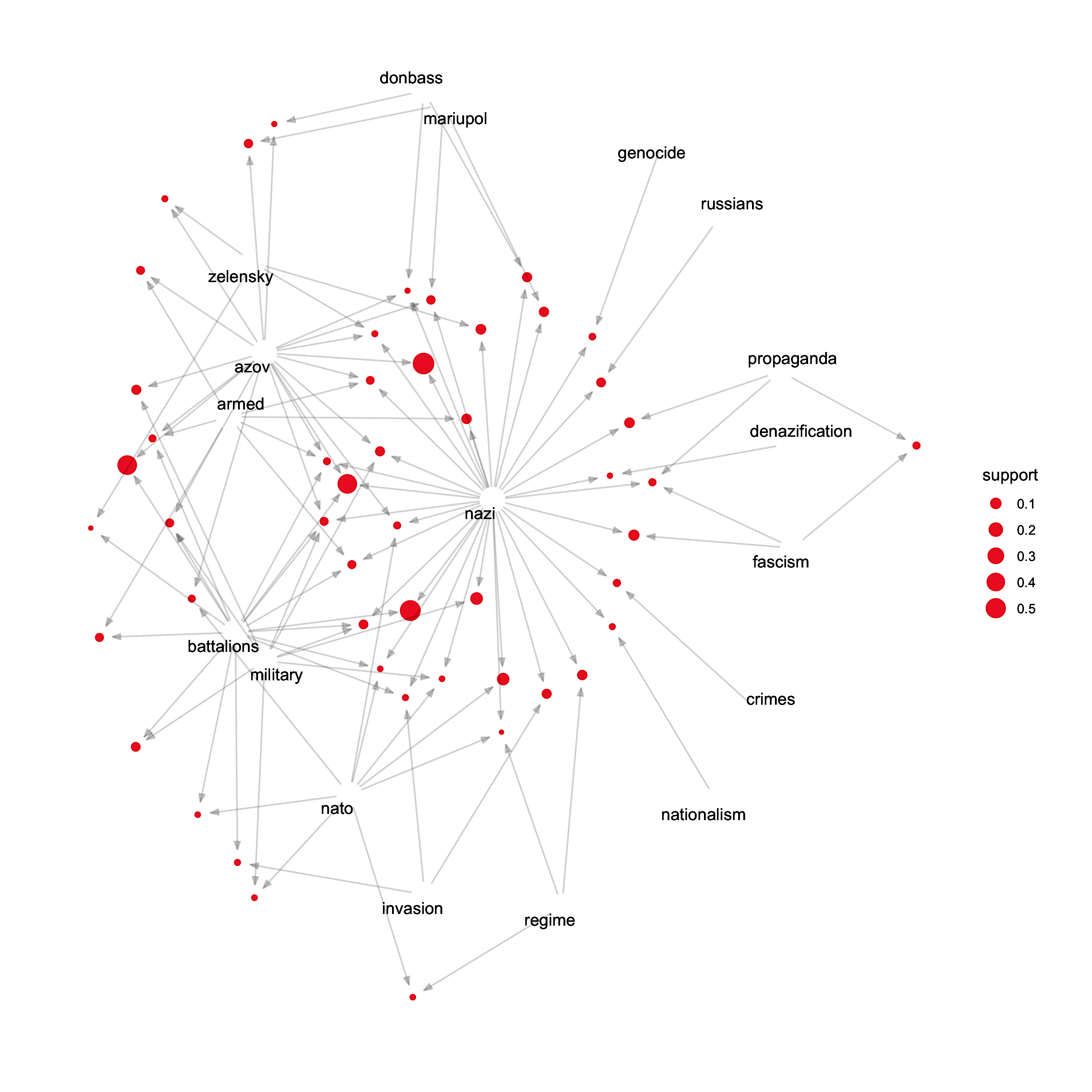}
 \caption{Graph of  semantic frequent itemsets}
 \label{fritemsets_nazi_1}
 \end{figure}

\begin{figure}
\center
 \includegraphics[width=0.95\linewidth]{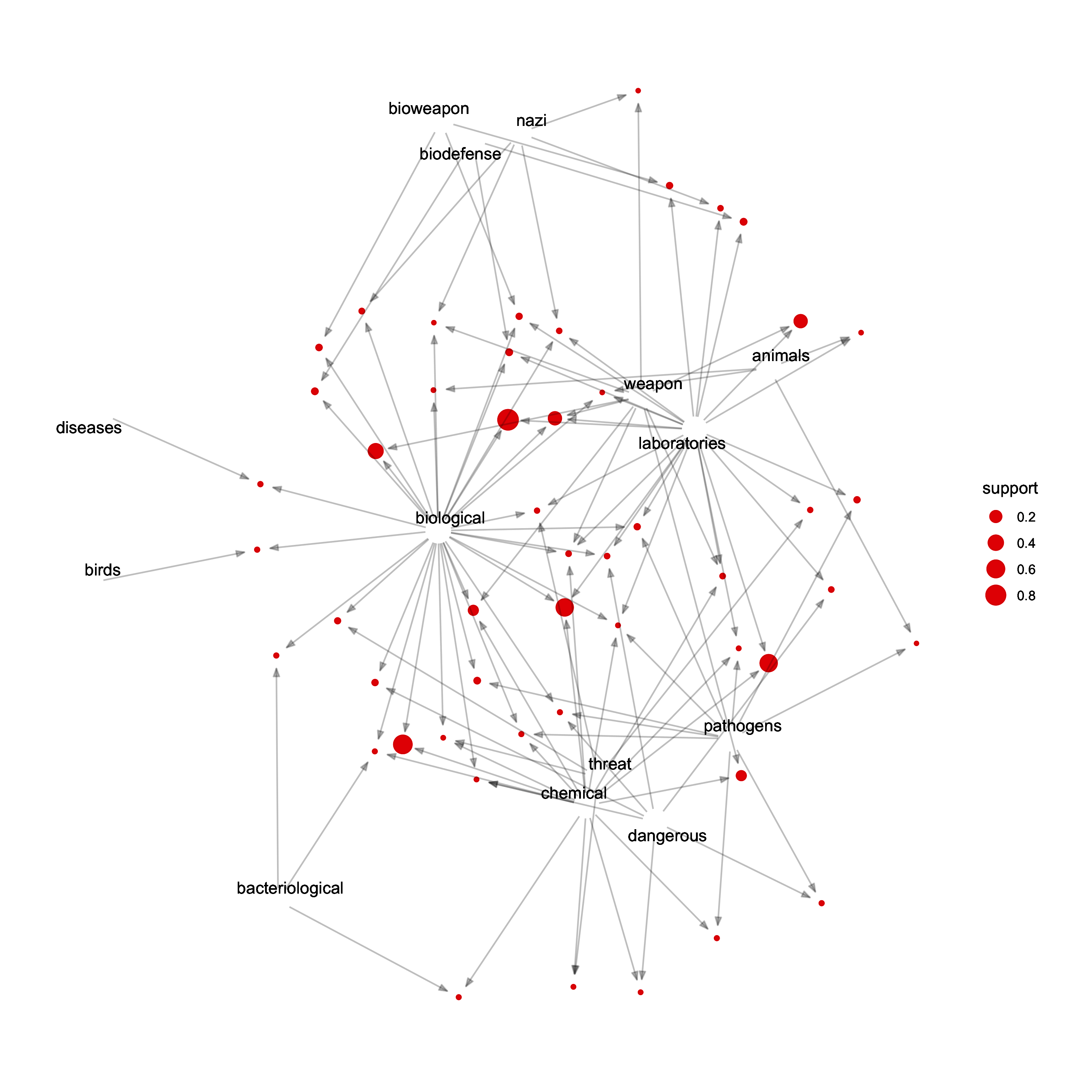}
 \caption{Graph of  semantic frequent itemsets.}
 \label{fritemsets_bioweapon_1}
 \end{figure}

In the paper~\cite{pavlyshenko2022methods}, we also consider the use of frequent itemsets and and associative rules for analysing tweets.
The frequent itemsets  and associative rules 
 can be used in a text data analysis to identify and analyze certain sets of objects, which are often found in large arrays and are characterized by certain features. 
Figure~\ref{fritemsets_nazi_1} shows the graph of frequent itemsets which describes the semantic structure of entities for a specified thematic field. 
Figure~\ref{fritemsets_bioweapon_1} shows similar calculation for the thematic field 'ukraine biological weapon'.  

\begin{figure}
\center
 \includegraphics[width=0.75\linewidth]{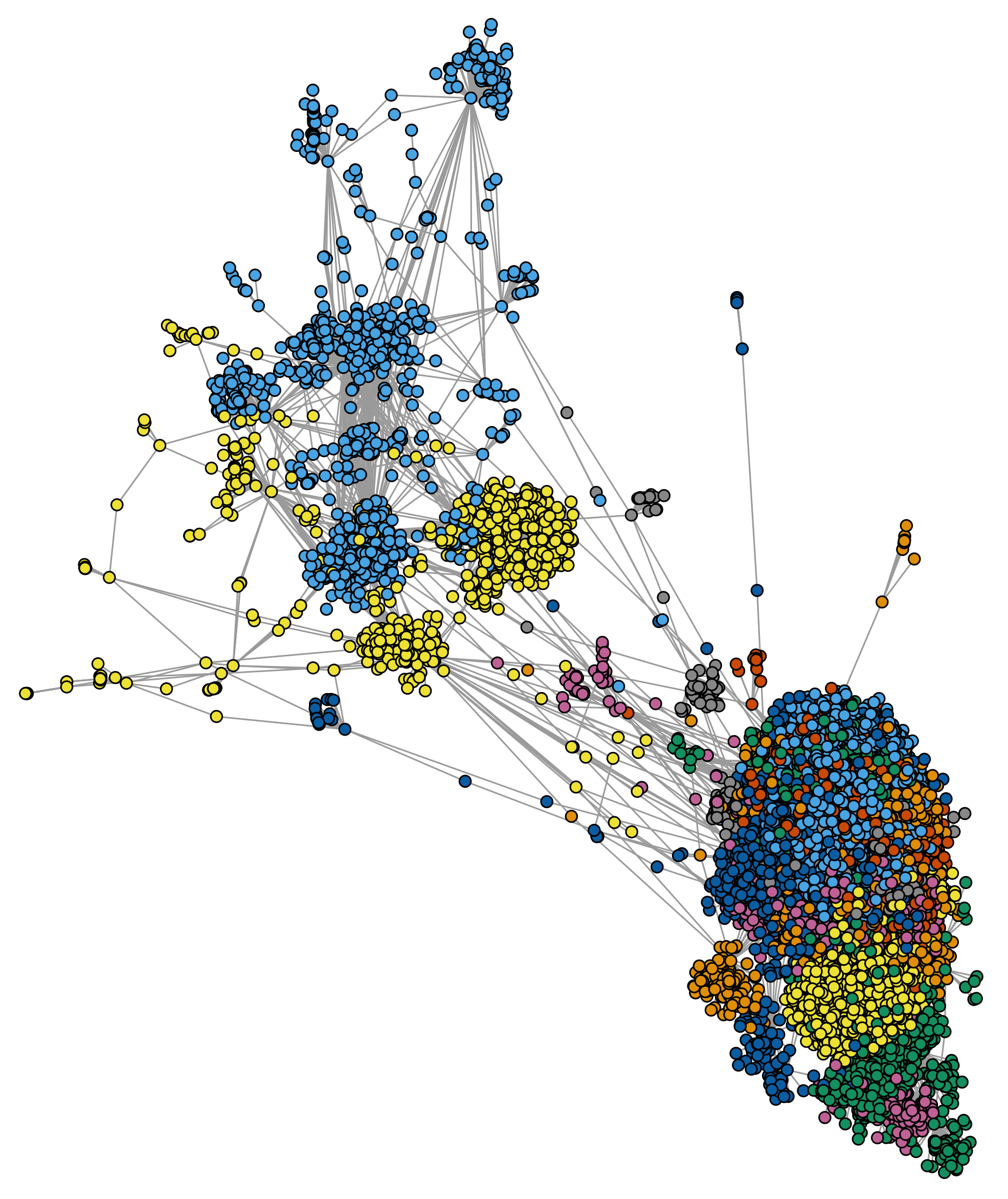}
 \caption{Graph of users' connections.}
 \label{usr_graph}
 \end{figure}

The  relationships among users can be considered as a graph, where vertices denote users and edges denote their connections.
Using graph mining algorithms, one can detect user communities and find ordered lists of users by various characteristics, such as
\textit {Hub, Authority, PageRank, Betweenness}. To identify  user communities, we used the \textit{Community Walktrap} algorithm and  to visualize them we used  \textit{Fruchterman-Reingold} algorithm, which are implemented in the package
 \textit{'igraph'}~\cite{csardi2006igraph} for the \textit{R} programming language environment. 
The \textit{Community Walktrap} algorithm searches for related subgraphs, also called communities, by random walk~\cite{pons2005computing}.
A graph which shows the relationships between users can be represented by
 Fruchterman-Reingold algorithm~\cite{fruchterman1991graph}.
 The qualitative structure of user's connections  can be used for aggregating different quantitative time series and, in such a way, creating new features for predictive models which can be used, for example, for predicting target variables.  
 Figure~\ref{usr_graph} shows users' connections and revealed communities for the subset of tweets which are related to the trends under consideration.  
 The results show that some communities marked by different colors are highly isolated and have only few connections outside.  This kind of communities can be treated as suspicious,  since artificially created communities for amplifying manipulative news are also highly isolated and their activity is often concentrated on amplification by retweeting tweets from a limited set of users. Therefore,  the numerical characteristics of users communities can have a predictive potential. 
 
\section{Parameter-efficient Fine-Tuning Llama 2 LLM}
Llama 2 model is considered in the work~\cite{touvron2023llama}  as 
 a collection of pretrained and fine-tuned large language models (LLMs) ranging in scale from 7 billion to 70 billion parameters. Fine-tuned LLMs, called Llama 2-Chat, are optimized for dialogue use cases. These models outperform open-source chat models on most benchmarks they have been tested, and based on  human evaluations for helpfulness and safety, they may be a suitable substitute for closed-source models. The paper~\cite{touvron2023llama} provides a detailed description of the approach to fine-tuning and safety improvements of Llama 2-Chat. 
Full fine-tuning is applicable in the case when we need to ingest millions of documents into LLM. But in the case of much smaller data, we can use a PEFT/LoRA approach which consists in fine-tuning a much smaller number of model parameters. These parameters are saved in the model adapter which is used for full model modification before using it for the model text response generation.  To optimize GPU usage, 
4bit or 8bit quantization of LLM can be chosen for model fine-tuning.
State-of-the-art Parameter-Efficient Fine-Tuning (PEFT) methods enable efficient adaptation of pre-trained language models (PLMs) to various downstream applications without fine-tuning all the model's parameters. Fine-tuning large-scale PLMs is often prohibitively costly. In this regard, PEFT methods only fine-tune a small number of (extra) model parameters, thereby greatly decreasing the computational and storage costs. Recent State-of-the-Art PEFT techniques achieve the performance comparable to that of full fine-tuning~\cite{peft}.
The paper~\cite{hu2021lora} considers Low-Rank Adaptation, or LoRA, which freezes pre-trained model weights and injects trainable rank decomposition matrices into each layer of the Transformer architecture, greatly reducing the number of trainable parameters for downstream tasks.  PEFT/LoRA approach approach makes it possible to fine-tune LLMs with sizes near 7B parameters, using Google Colab.  Along with the text data for fine-tuning, it is important to use prompt instructions which show how to process input prompts. Instructions can be created by human experts and augmented by other LLM models. 
LLM generates complex output texts on prompts which can be optimized in different ways. One possible way is selecting appropriate instructions for fine-tuning models. Another way is using a method called Reinforcement Learning from Human Feedback (RLHF)~\cite{beeching2023stackllama}. In this approach, human experts estimate and rate LLM output and then using this rates as target lables, LLM can be fine-tuned by supervised training.  

Let us consider fine-tuning Llama 2 using a PEFT/LoRA approach. 
The instruction for fine-tuning can be created in different ways, e.g. by experts or using LLMs like ChatGPT, GPT-4, Llama with appropriate prompts which specify LLM response.
 The dataset of fake news was taken from Kaggle repository~\cite{kagglefakenews, ahmed2018detecting}. 
 To the training datasets, we also added Russian propaganda narratives which were taken from Vox Ukraine~\cite{russnarratives}.
 For testing the fine-tuning, we used the dataset with prompt instructions for analysing a text, 
highlighting the main points of a text, summarizing a text and extracting named entities with appropriate sentiments.  The dataset was split into training and validation datasets, the relative part of validation dataset was 0.25.
The approaches for using PEFT/LoRA for fine-tuning Llama 2 are described in~\cite{hfllama2, vonwerra2022trl}.
 For the model fine-tuning the trainer \verb|SFTTrainer| 
from package \verb|trl|~\cite{vonwerra2022trl} was used.
The following training arguments were set up:
\begin{newmargin}{1cm}{0cm} 
\begin{verbatim}
'model_name':'meta-llama/Llama-2-7b-chat-hf',
learning_rate':5e-4, 
'num_train_epochs':10,
'max_seq_length':2048,
'gradient_accumulation_steps':2, 
'load_in_4bit': True,
'bnb_4bit_quant_type':'nf4',
'lr_scheduler_type':'linear'
\end{verbatim}
\end{newmargin}

\FloatBarrier

\section{Testing Fine-Tuned Llama 2 Model}
Let us consider the results of testing the fine-tuned Llama 2 model. 
The prompt for analysing the text using fine-tuned LLama 2 LLM has the following structure:
\begin{newmargin}{0.5cm}{0cm} 
\begin{verbatim}
<s>[INST] <<SYS>>
You are an expert in fake news and disinformation revealing 
and analytics.
<</SYS>>
      
{ prompt } [/INST]
\end{verbatim}
\end{newmargin}
where \verb|{ prompt }| -- is the prompt which consists of a question and text for the analysis. For example, the following question can be used for text analytics: 'Please analyse the following text from the perspective of revealing disinformation and manipulation'.
We tested Llama 2 LLM model on  the analysis of an input text with the following tasks: analysing a text on revealing disinformation and  propaganda narratives, fact checking, fake news detection,  manipulation analytics, extracting named entities with their sentiments. 
For the analysis, we have chosen several recent  statements, claims and thesis of well-known people concerning Russian war in Ukraine. 

Hungarian Prime minister Victor Orban gave an interview to Tacker Carlson~\cite{politicoorbalcarlson}. The following are the prompt based on thesis and statements of this interview and the response of fine-tuned Llama 2 model:
\begin{newmargin}{0.5cm}{0cm} 
{\fontsize{9}{9} \fontfamily{lmtt} \selectfont 
Prompt: 
 \\  
 \\  {<}s{>}{[}INST{]} {<}{<}SYS{>}{>} 
 \\ You are an expert in fake news and disinformation revealing and analytics. 
 \\ {<}{<}/SYS{>}{>} 
 \\  
 \\ Please analyse input text from perspective of revealing disinformation narratives. 
 \\ Input text: 
 \\ Hoping Ukraine will beat Russia "is not just a misunderstanding. It is a lie. It’s impossible. 
  Ukraine has no chance of winning the war against Russia. 
  We missed the historic opportunity” to admit Ukraine to NATO, Russia is getting "stronger and stronger."  
  Ukraine’s admission to NATO "is not a realistic proposal at this moment, so forget about it". 
 \\ "Call back Trump! Because you know, you can criticize him for many reasons … but … the best foreign policy of the recent several decades belongs to him. He did not initiate any new war, he treated nicely the North Koreans, and Russia and even the Chinese … and if he would have been the president at the moment of the Russian invasion {[}of Ukraine{]}, it would be not possible to do that by the Russians." 
 \\ "Trump is the man who can save the Western world" and all of humanity. {[}/INST{]}  
 \\  
 \\Response: 
 \\  
 \\ The first narrative in the input text is that Ukraine cannot win the war against Russia. This narrative is manipulative as it suggests that Ukraine is incapable of success, which is a subjective claim not based on factual evidence.  
 \\ The second narrative suggests that Russia missed a historic opportunity to admit Ukraine to NATO. This narrative is misleading as it oversimplifies the complex geopolitical situation and ignores the fact that Ukraine's admission to NATO requires a consensus from all member states, which is difficult to achieve due to various reasons. 
 \\ The third narrative suggests that the US President, Donald Trump, is the only savior who can save the Western world and humanity from Russia. This is manipulative as it uses an emotional appeal to sway readers' opinion, ignoring the fact that Trump's policies are controversial and have been criticized by many. 
 \\ The offensive lexicon used in the text includes terms like 'lie','missed','stronger', 'foreign policy', 'initiate', 'treated','man who can save the Western world', 'humanity'. These words are used to create a negative sentiment towards Ukraine, NATO and the US, while glorifying Russia.  
 \\ The named entities associated with negative sentiment in the text include Ukraine, NATO, and Trump, while Russia is associated with a positive sentiment.   
 \\ The overall tone of the text is manipulative, using emotional language and subjective claims to shape readers' opinion. It manipulates the readers' perceptions by presenting a one-sided view of the situation, which is not based on factual evidence.
 }
\end{newmargin}

The following are the prompt and  the fine-tuned Llama 2 model response for Tucker Carlson's claims and statements on the support for Ukraine~\cite{voxcheckcarlsontaker}:
\begin{newmargin}{0.5cm}{0cm} 
{\fontsize{9}{9} \fontfamily{lmtt} \selectfont 
 Prompt: 
 \\  
 \\  {<}s{>}{[}INST{]} {<}{<}SYS{>}{>} 
 \\ You are an expert in fake news and disinformation revealing and analytics. 
 \\ {<}{<}/SYS{>}{>} 
 \\  
 \\ Please analyse input text from perspective of revealing disinformation narratives and sarcasm. 
 \\ Input text: 
 \\ See, it’s very easy to understand. It is vitally important for you to support Ukraine because it’s necessary for Ukraine to be supported by you. Your support is mandatory until it’s finished, whatever it is and whatever that means. So shut up and support Ukraine or else you’re in trouble. Back when they still taught logic, statements like this were known as tautologies. Something is true because it is. The more you repeat it, the truer it becomes. It’s a self-reinforcing reality. There was a time when tautologies were considered illegitimate arguments, not to mention hilariously stupid. {[}/INST{]}  
 \\  
 \\ Response: 
 \\  
 \\ 1. The lexicon used is offensive and sarcastic, targeting those who support Ukraine. Phrases like'shut up and support Ukraine or else you’re in trouble' and 'tautologies were considered illegitimate arguments' are clearly intended to belittle and mock those who disagree with the author's stance.  
 \\ 2. The narrative presented is manipulative, suggesting that supporting Ukraine is a compulsory action, without providing any factual basis for this claim. This is misleading and manipulative, as it implies that not supporting Ukraine could lead to negative consequences without providing any evidence.  
 \\ 3. The statement 'your support is mandatory until it’s finished, whatever it is and whatever that means' is misleading and ambiguous. It implies that there is no end-point for supporting Ukraine, which is not a fact-based statement.   
 \\ 4. The language used is emotional manipulation, aiming to provoke an emotional response from the reader.   
 \\ 5. The claim that tautologies were 'considered illegitimate arguments' is misleading. While tautologies are not considered legitimate arguments, they are not necessarily hilariously stupid. This statement is offensive and intended to belittle the idea of supporting Ukraine.   
 \\ 6. The essay uses sarcasm and mocking language to undermine the idea of supporting Ukraine. It mocks the idea of tautologies as a basis for argument, which is a clear sign of disinformation and manipulative rhetoric. 
 \\ In conclusion, the essay is heavily loaded with manipulative and offensive language, sarcasm, and disinformation. It contains several narratives that are not supported by factual evidence, and uses emotional manipulation to provoke an emotional response from the reader.
 }
\end{newmargin}

In the training dataset, we also added the information from Marc Thiessen's article  “America First” Case for Supporting Ukraine~\cite{thiessen2023america}. The following is the response of fine-tuned Llama 2 model to the question:"What are the critical reasons for the United States to maintain its support for Ukraine? ": 

\begin{newmargin}{0.5cm}{0cm} 
{\fontsize{9}{9} \fontfamily{lmtt} \selectfont 
 The United States has been a crucial supporter of Ukraine since the early 2010s, providing financial, military, and diplomatic support. Here are some critical reasons why the United States should maintain its support for Ukraine: 
 \\ 1. Geopolitical Stability: Ukraine is a key player in maintaining regional stability. The conflict in Ukraine has already led to a significant destabilization of the European region, and a failure to support Ukraine could lead to further escalations. The United States should continue to support Ukraine to prevent a wider conflict and maintain the stability of the region. 
 \\ 2. Countering Russian Aggression: The conflict in Ukraine is a direct result of Russian aggression. By supporting Ukraine, the United States is directly countering Russian aggression. Continued support will not only deter Russia from further aggressive actions but also prevent the spread of its influence in the region.  
 \\ 3. Protecting NATO Interests: Ukraine is a crucial buffer zone between Russia and NATO member states. The conflict in Ukraine is not only a matter of national security for Ukraine but also a matter of national security for NATO countries. Continued support will help maintain the security of NATO member states.  
 \\ 4. Supporting Democratic Values: The conflict in Ukraine is not only about geopolitics but also about the fight against authoritarianism and the defense of democratic values. The United States should continue to support Ukraine to uphold democratic principles and counter the influence of authoritarian regimes. 
 \\ 5. Promoting Economic Growth: The conflict in Ukraine has led to significant economic losses, particularly in the eastern regions. The United States should continue to support Ukraine to rebuild and restore its economy.  
 \\ 6. Supporting the Rule of Law: The conflict in Ukraine has also led to significant challenges to the rule of law. The United States should continue to support Ukraine to strengthen its judicial system and fight against corruption. 
 \\ 7. Strengthening International Order: The conflict in Ukraine has challenged the international order, particularly the principles of sovereignty and territorial integrity. The United States should continue to support Ukraine to uphold these principles and strengthen the international order.  
 \\ 8. Protecting Humanitarian Interests: The conflict in Ukraine has led to significant humanitarian crises, particularly in the eastern regions. The United States should continue to support Ukraine to address these crises and protect the human rights of its citizens. 
 \\ In conclusion, the United States has significant reasons to continue its support for Ukraine. It is essential to maintain stability in the region, counter Russian aggression, protect NATO interests, support democratic values, promote economic growth, strengthen the rule of law, and uphold the international order. Moreover, supporting Ukraine is crucial to protecting humanitarian interests and addressing significant humanitarian crises. 
}
\end{newmargin}

The following are the prompt and  fine-tuned Llama 2 for fact checking about  destroying Kakhovka Dam in Tucker Carlson's Twitter video episode ~\cite{tuckercarlsonfactcheck,voxtackercarlson} using 
claims and statements from this video.

\begin{newmargin}{0.5cm}{0cm} 
{\fontsize{9}{9} \fontfamily{lmtt} \selectfont 
 \textbf{Prompt:} 
 \\ {<}s{>}{[}INST{]} {<}{<}SYS{>}{>} 
 \\ You are an expert in fake news and disinformation revealing and analytics. 
 \\ {<}{<}/SYS{>}{>} 
 \\  
 \\ Please analyse input text from perspective of revealing disinformation narratives. 
 \\ Input text: 
 \\ The question is, who did it? Well, let’s see. The Kakhovka Dam was effectively Russian. It was built by the Russian government. It currently sits in Russian-controlled territory. The dam’s reservoir supplies water to Crimea, which has been for the last 240 years home of the Russian Black Sea Fleet. Blowing up the dam may be bad for Ukraine, but it hurts Russia more.  
 \\ And for precisely that reason, the Ukrainian government has considered destroying it. 
 \\ Write finish conclusion if we can treat this text as fake news. 
 \\   {[}/INST{]} 
 \\  
 \\ \textbf{Response:} 
 \\  
 \\ Analyzing the input text, several narratives can be identified. The first narrative is that the Kakhovka Dam is Russian, a claim that is false. The dam was built by the Soviet government, not the Russian government, and currently sits in Ukrainian territory. The second narrative is that the dam supplies water to Crimea, which is a misleading statement as Crimea is landlocked. The third narrative is that the Ukrainian government considered destroying the dam, which is not supported by any credible sources. 
 \\ The text also uses manipulative and emotional language, such as 'bad for Ukraine, but it hurts Russia more,' which is subjective and designed to elicit an emotional response from the reader. The text also uses a rhetorical device, 'the Ukrainian government has considered destroying it,' which is not supported by any credible sources. 
 \\ In conclusion, this text can be treated as fake news. It contains misinformation and manipulative content, designed to promote a narrative that is not supported by factual evidence. 
 \\  
 }
\end{newmargin}

Bulgarian President Rumen Radev told the reporters on Friday that he wanted “to make it clear that Ukraine insists on fighting this war,” the Associated Press reported. “But it should also be clear that the bill is paid by the whole of Europe,” he said.~\cite{ukrblastbulgpresident}.
The following are the prompt and  fine-tuned Llama 2 for the analysis of this statement. 
\begin{newmargin}{0.5cm}{0cm} 
{\fontsize{9}{9} \fontfamily{lmtt} \selectfont 
Prompt: 
 \\  
 \\  {<}s{>}{[}INST{]} {<}{<}SYS{>}{>} 
 \\ You are an expert in fake news and disinformation revealing and analytics. 
 \\ {<}{<}/SYS{>}{>} 
 \\  
 \\ Please analyse the following text from perspective of revealing disinformation narratives: 
 \\ Ukraine insists on carrying on with this war, but the whole of Europe is paying the bills.  {[}/INST{]}  
 \\  
 \\ Response: 
 \\  
 \\ 1. The narrative that Ukraine is responsible for the war. This narrative is manipulative as it implies that Ukraine is the aggressor, without providing any evidence to support this claim.  
 \\ 2. The narrative that Europe is paying for Ukraine's war. This is misleading. While it is true that the conflict has affected European countries, the claim that Europe is directly financing the war is not supported by any credible evidence. 
 \\ 3. The narrative that the West is profiting from the war. This is manipulative and lacks factual basis. It implies that the West is exploiting the conflict for its own gains, which is not supported by any reliable sources. 
 \\ 4. The lexicon used in the text is offensive and emotional, aiming to provoke an emotional response from the reader. 
 \\ The named entities involved in this narrative include 'Ukraine' and 'Europe'. The entities are presented in a negative light, suggesting that Ukraine is the aggressor and Europe is being burdened financially. 
 \\ The claim that Ukraine is responsible for the war is a subjective and controversial claim that lacks a factual basis. The claim that Europe is paying for the war is misleading and not supported by any credible evidence. The claim that the West is profiting from the war is manipulative and lacks factual basis. 
 \\ The text uses emotional language and manipulative content to create a negative perception of Ukraine and Europe, while the West is portrayed in a positive light. However, these narratives are not based on factual information and are instead designed to manipulate public opinion.
 }
\end{newmargin}

According to the Hungarian official, to achieve lasting peace, "the Western world, which supports Ukraine, must provide security guarantees to Russia, but not NATO membership for Ukrainians."~\cite{ungarianofficialspeaks}.
The following is the analysis  of this thesis by the fine-tuned Llama 2 model.

\begin{newmargin}{0.5cm}{0cm} 
{\fontsize{9}{9} \fontfamily{lmtt} \selectfont 
 Prompt:
  \\ 
 \\ {<}s{>}{[}INST{]} {<}{<}SYS{>}{>} 
 \\ You are an expert in fake news and disinformation revealing and analytics. 
 \\ {<}{<}/SYS{>}{>} 
 \\ Please analyse input text from perspective of revealing disinformation narratives. 
 \\ Input text: 
 \\ According to the Hungarian official,  
 \\ to achieve lasting peace, "the Western world, which supports Ukraine, must provide  
 \\ security guarantees to Russia, but not NATO membership for Ukrainians {[}/INST{]} 
  \\ 
  \\ Response:
\\
 \\ Analysis:  
 \\ The input text contains several narratives that are often associated with disinformation and fake news. 
 \\ 1. "The Western world, which supports Ukraine, must provide security guarantees to Russia": This is a manipulative narrative. It suggests that the West's support for Ukraine is a threat to Russia. This is a classic example of a 'chicken game', where the West is portrayed as threatening Russia, which is not supported by factual evidence. 
 \\ 2. "but not NATO membership for Ukrainians": This is a controversial claim. Ukraine has been a member of NATO since 2014, which is not a factual statement. It is also misleading as it implies that Ukraine joining NATO is a threat to Russia, which is not a valid argument. 
 \\ 3. "the Western world, which supports Ukraine, must provide security guarantees to Russia": This statement is manipulative and contains a lack of factual basis. It is presented as if it is a universally accepted truth, which it is not. 
 \\ 4. Contains offensive and emotional language: The use of terms such as "Hungarian official" and "Western world" can be seen as offensive and manipulative.  
 \\ 5. Contains a personal attack: The text implies that the 'Hungarian official' is manipulating the situation.  
 \\ Conclusion: 
 \\ The input text contains disinformation narratives, manipulative language, and lacks factual basis. It is not a reliable source of information. The narratives presented are designed to mislead the reader and create a negative perception of the West and Ukraine.
}
\end{newmargin}

We also conducted the testing of the fine-tuned Llama 2 model on analysing and extracting named entities with appropriate sentiments and expalnations using arbitrary news from the dataset~\cite{kagglefakenews, ahmed2018detecting}. The results 
of testing and extracting entities with coresponding sentiments are given in the Appendix.
\section{Conclusion}
 The main goal of the study is to see if the analytical capability of the fine-tuned LLM has a predictive potential and can generate a response with a specified structure which can be useful in disinformation analytics and fake news detection.
  
 For fine-tuning, the PEFT/LoRA based approach was used. The model was fine-tuned for the following tasks: analysing a text on revealing disinformation and  propaganda narratives, fact checking, fake news detection,  manipulation analytics, extracting named entities with their sentiments. The PEFT/LoRA approach makes it possible to use cheap GPU resources for model fine-tuning. 
The obtained results show that fine-tuned Llama 2 model can perform multitask news analyses with a specified structure of response, part of response can be a structured text and another part of data can have a JSON format that is convenient for further processing of LLM response. 
The PEFT/LoRA approach makes it possible to use cheap GPU resources for model fine-tuning which are available, e.g. on Google Colab  in case of small input text size and  4bit quantization. Taking into account that LLM model can generate the output in specified JSON format, the data of sentiments for named entities can be used in predictive models as features and can be loaded directly into predictive models via appropriate API.  These features can have a predictive potential for different target variables including quantitative characteristics  of companies' behavior on financial markets. 

The obtained results on fine-tuning  Llama2 LLM model using PEFT/LoRA approah can be assessed  qualitatively by experts to see if the news analytics, summarizing, highlighting main points and entities extracting give some essential information to news experts.  
The considered approach can show high efficiency, using small sets of instructions due to the LLM ability of few-shot learning that is not inherent for conventional transformer based models. To improve the LLM performance, one needs a more precisely created training dataset and to exploit the RLHF method to get a more optimized LLM. The considered approach can be applied for using extracted entities and sentiments in supervised models with quantitative target variables, e.g. for the analysis of companies' behavior on financial and business markets. 

The fine-tuned Llama 2 model was tested on different news data including thesis, statements  and claims of known persons. The results show 
that a fine-tuned model can conduct inferences and give chain of thoughts regarding the text under analysis. 
Quality and usefulness of the analysis by LLM depend on the quality of training datasets.  In the test results, one can mention some inaccuracy. This problem can be fixed using more accurate and specified training dataset with the training dataset with instructions corrected  by an expert. 
 In LLM responses, we can find that some generated analytical points are not precise or informative. For testing, we used a very small training dataset. To make responses more presise and informative, one can use a larger dataset selected by experts using the methods of active learning with iterative adding of new additional data which describe problematic topics and points.  For testing, we used a small datatset with the 4bit model quantization for small Llama2 model with 7B parameters. The results show that with a larger dataset, larger Llama 2 model( e.g. with 13B or 70B parameters) and with 8bit or 16bit parameters quantization, one can expect an essential improvement of the quality of the analysis conducted by the fine-tuned Llama 2 model. 

The obtained results show that th fine-tuned Llama 2 LLM model can be used as a part of complex deep learning approach for the analysis of news from the perspective of detecting disinformation, propaganda, fake news and manipulations. 

\section{Disclaimer} 
We are sharing a considered approach, ideas and results for academic purpose only, not for any real conclusions or recommendations.

\bibliographystyle{unsrt}
\bibliography{article.bib}

\begin{thebibliography}{10}

\bibitem{pavlyshenko2022methods}
Bohdan~M. Pavlyshenko.
\newblock {Methods of Informational Trends Analytics and Fake News Detection on
  Twitter}.
\newblock {\em arXiv preprint arXiv:2204.04891, Download PDF:
  https://arxiv.org/pdf/2204.04891.pdf}, 2022.

\bibitem{pavlyshenko2023financial}
Bohdan~M Pavlyshenko.
\newblock Financial news analytics using fine-tuned llama 2 gpt model.
\newblock {\em arXiv preprint arXiv:2308.13032}, 2023.

\bibitem{touvron2023llama}
Hugo Touvron, Louis Martin, Kevin Stone, Peter Albert, Amjad Almahairi, Yasmine
  Babaei, Nikolay Bashlykov, Soumya Batra, Prajjwal Bhargava, Shruti Bhosale,
  et~al.
\newblock {Llama 2: Open foundation and fine-tuned chat models}.
\newblock {\em arXiv preprint arXiv:2307.09288}, 2023.

\bibitem{grinberg2019fake}
Nir Grinberg, Kenneth Joseph, Lisa Friedland, Briony Swire-Thompson, and David
  Lazer.
\newblock Fake news on twitter during the 2016 us presidential election.
\newblock {\em Science}, 363(6425):374--378, 2019.

\bibitem{ajao2018fake}
Oluwaseun Ajao, Deepayan Bhowmik, and Shahrzad Zargari.
\newblock Fake news identification on twitter with hybrid cnn and rnn models.
\newblock In {\em Proceedings of the 9th international conference on social
  media and society}, pages 226--230, 2018.

\bibitem{helmstetter2018weakly}
Stefan Helmstetter and Heiko Paulheim.
\newblock Weakly supervised learning for fake news detection on twitter.
\newblock In {\em 2018 IEEE/ACM International Conference on Advances in Social
  Networks Analysis and Mining (ASONAM)}, pages 274--277. IEEE, 2018.

\bibitem{pavlyshenko2019forecasting}
Bohdan~M. Pavlyshenko.
\newblock {Forecasting of Events by Tweets Data Mining}.
\newblock {\em Electronics and information technologies}, (10):71--85, 2018.

\bibitem{pavlyshenko2019cantwitter}
Bohdan~M. Pavlyshenko.
\newblock {Can Twitter Predict Royal Baby's Name ?}
\newblock {\em Electronics and information technologies}, (11):52--60, 2019.

\bibitem{pavlyshenko2021forming}
Bohdan~M Pavlyshenko.
\newblock Forming predictive features of tweets for decision-making support.
\newblock In {\em International Scientific Conference “Intellectual Systems
  of Decision Making and Problem of Computational Intelligence”}, pages
  479--490. Springer, 2021.

\bibitem{pavlyshenko2022forming}
Bohdan~M. Pavlyshenko.
\newblock {Forming Predictive Features of Tweets for Decision-Making Support}.
\newblock In {\em Lecture Notes in Computational Intelligence and Decision
  Making: 2021 International Scientific Conference" Intellectual Systems of
  Decision-making and Problems of Computational Intelligence”, Proceedings},
  pages 479--490. Springer, Download PDF: https://arxiv.org/pdf/2201.02049,
  2022.

\bibitem{csardi2006igraph}
Gabor Csardi, Tamas Nepusz, et~al.
\newblock The igraph software package for complex network research.
\newblock {\em InterJournal, complex systems}, 1695(5):1--9, 2006.

\bibitem{pons2005computing}
Pascal Pons and Matthieu Latapy.
\newblock Computing communities in large networks using random walks.
\newblock In {\em International symposium on computer and information
  sciences}, pages 284--293. Springer, 2005.

\bibitem{fruchterman1991graph}
Thomas~MJ Fruchterman and Edward~M Reingold.
\newblock Graph drawing by force-directed placement.
\newblock {\em Software: Practice and experience}, 21(11):1129--1164, 1991.

\bibitem{peft}
Sourab Mangrulkar, Sylvain Gugger, Lysandre Debut, Younes Belkada, and Sayak
  Paul.
\newblock {PEFT: State-of-the-art Parameter-Efficient Fine-Tuning methods}.
\newblock \url{https://github.com/huggingface/peft}, 2022.

\bibitem{hu2021lora}
Edward~J Hu, Yelong Shen, Phillip Wallis, Zeyuan Allen-Zhu, Yuanzhi Li, Shean
  Wang, Lu~Wang, and Weizhu Chen.
\newblock {Lora: Low-rank adaptation of large language models}.
\newblock {\em arXiv preprint arXiv:2106.09685}, 2021.

\bibitem{beeching2023stackllama}
Edward Beeching, Younes Belkada, Kashif Rasul, Lewis Tunstall, Leandro von
  Werra, Nazneen Rajani, and Nathan Lambert.
\newblock Stackllama: A hands-on guide to train llama with rlhf.
\newblock \url{https://huggingface.co/blog/stackllama }, 2023.

\bibitem{kagglefakenews}
{ Fake and real news dataset. Kaggle.Com}.
\newblock
  \url{https://www.kaggle.com/datasets/clmentbisaillon/fake-and-real-news-dataset
  }.

\bibitem{ahmed2018detecting}
Hadeer Ahmed, Issa Traore, and Sherif Saad.
\newblock Detecting opinion spams and fake news using text classification.
\newblock {\em Security and Privacy}, 1(1):e9, 2018.

\bibitem{russnarratives}
{ Propaganda Diary. Vox Ukraine }.
\newblock \url{https://russiandisinfo.voxukraine.org/en/narratives}.

\bibitem{hfllama2}
HuggingFace.
\newblock Llama 2 is here - get it on hugging face.
\newblock \url{https://huggingface.co/blog/llama2}, 2023.

\bibitem{vonwerra2022trl}
Leandro von Werra, Younes Belkada, Lewis Tunstall, Edward Beeching, Tristan
  Thrush, Nathan Lambert, and Shengyi Huang.
\newblock {TRL: Transformer Reinforcement Learning}.
\newblock \url{https://github.com/huggingface/trl}, 2020.

\bibitem{politicoorbalcarlson}
{Viktor Orbán tells Tucker Carlson: Trump’s the man to save the West}.
\newblock
  \url{https://www.politico.eu/article/hungarian-pm-viktor-orban-donald-trump-ukraine-victory-over-russia-lie/}.

\bibitem{voxcheckcarlsontaker}
Sholudko V., Brodovska A., and Tkachenko Y.
\newblock {The voice of propaganda in the USA is returning. Fact-check of the
  first episode of the Tucker Carlson show. Vox Ukraine.}
\newblock
  https://voxukraine.org/en/the-voice-of-propaganda-in-the-usa-is-returning-
  fact-check-of-the-first-episode-of-the-tucker-carlson-show.

\bibitem{thiessen2023america}
{Marc A. Thiessen. The “America First” Case for Supporting Ukraine. The
  Washington Post }.
\newblock
  \url{https://www.washingtonpost.com/opinions/2023/05/30/ukraine-counteroffensive-support-america}.

\bibitem{tuckercarlsonfactcheck}
{Tom Norton. Fact Check: Tucker Carlson Says Ukraine Considered Destroying
  Kakhovka Dam. Newsweek.}
\newblock
  \url{https://www.newsweek.com/fact-check-tucker-carlson-ukraine-destroyed-kakhovka-dam-claim-1805277},
  2023.

\bibitem{voxtackercarlson}
{ The voice of propaganda in the USA is returning. Fact-check of the first
  episode of the Tucker Carlson show. Vox Ukraine}.
\newblock
  https://voxukraine.org/en/the-voice-of-propaganda-in-the-usa-is-returning-fact-check-of-the-first-episode-of-the-tucker-carlson-show.

\bibitem{ukrblastbulgpresident}
{Ukraine blasts Bulgaria president’s claims that Kyiv is to blame for the
  war}.
\newblock
  \url{https://www.politico.eu/article/rumen-radev-bulgaria-president-ukraine-war/}.

\bibitem{ungarianofficialspeaks}
{Hungarian official speaks of "security guarantees" for Russia and hindering
  Ukraine's ascension to NATO}.
\newblock \url{https://www.pravda.com.ua/eng/news/2023/09/3/7418249/}.

\end{thebibliography}
\FloatBarrier
\begin{appendices}
\section{Analysis of Fake News And Extracting Named Entities}
For testing the fine-tuned Llama 2 LLM model, we took fake news from Kaggle Fake and Real News Dataset~\cite{kagglefakenews, ahmed2018detecting}. 
\subsection{Analysis of  News}
For our analysis, we use the following prompt  for the fine-tuned Llama 2  LLM model:

\begin{newmargin}{1cm}{0cm} 
\begin{verbatim}
<s>[INST] <<SYS>>
You are an expert in fake news and disinformation revealing and analytics.
<</SYS>>
      
Please analyse the following text from perspective of revealing 
disinformation and manipulation. 
 { Input text: } [/INST]
\end{verbatim}
 \end{newmargin}
 The following are input texts and responses of the fine-tuned Llama 2 model.
\begin{newmargin}{0.5cm}{0cm} 
{\fontsize{9}{9} \fontfamily{lmtt} \selectfont 
 \textbf{Input text:}
 \\ FBI DIRECTOR CONFIRMS Hillary’s Worst Nightmare Was Found On This Creep’s Laptop {[}VIDEO{]}. Be careful what you ask for Watch:  FBI Director James Comey testified Wednesday before the Senate Judiciary Committee that Hillary Clinton emails containing classified information were forwarded to the computer of Anthony Weiner, the former congressman, by his wife, Clinton aide Huma Abedin.Comey also strongly defended himself on Capitol Hill for his Oct. 28 letter notifying Congress of newly discovered emails appearing pertinent to Clinton s private server investigation, arguing it would have been  catastrophic  to conceal such a revelation.He explained that his investigative team informed him in late October of metadata from Weiner s computer showing thousands of Clinton s emails on that device, including what they thought might be the  missing emails  from her first three months at the State Department.Weiner resigned from Congress in 2011 over a sexting scandal and saw a 2013 mayoral run in New York City go up in flames for similar reasons. The investigation that led to Comey s letter concerned his sexting with an underage girl.  WFB 
 \\  
 \\ \textbf{Response:} 
 \\  
 \\ The text contains several controversial claims, which can be interpreted as misinformation, and some points of the text can be considered as manipulative news. 
 \\ 1. Controversial Claim: The text implies that Hillary Clinton's worst nightmare was found on this person's laptop, which is a subjective claim and lacks a factual basis. It is a personal attack on Hillary Clinton and is used to provoke emotional response from readers.  
 \\ 2. Controversial Claim: The text states that Hillary Clinton's emails containing classified information were forwarded to the computer of Anthony Weiner by his wife, Clinton aide Huma Abedin. The claim that Hillary Clinton forwarded classified emails to Anthony Weiner's laptop is controversial and lacks direct evidence. It is also presented in a sensationalist way to generate a negative reaction from readers. 
 \\ 3. Manipulative News: The text uses loaded language and emotional appeal to manipulate readers' emotions. Words such as "creep," "worst nightmare," and "sexting scandal" are used to create a negative image of the individuals involved. 
 \\ 4. Lack of Factual Basis: The text claims that the FBI Director James Comey "strongly defended himself on Capitol Hill for his Oct. 28 letter notifying Congress of newly discovered emails appearing pertinent to Clinton's private server investigation," but it does not provide any evidence to support this claim. 
 \\ 5. Misleading Content: The text states that Comey "argued it would have been 'catastrophic' to conceal such a revelation," but it does not provide context or further details about this statement, which can be misleading.  
 \\ In conclusion, the text contains several controversial claims, manipulative language, and lacks factual basis, which can be seen as misinformation or manipulative news. It is essential to cross-check such information with reputable sources before forming opinions or making decisions based on it. 
 \\  
  \\  
 ------------------------------------------------------------------------------------
 \\  
 \\ \textbf{Input text:} 
 \\ CHELSEA CLINTON Confronted by Woman at Book Signing: “Can you sign this for my friend Juanita Broaddrick” {[}Video{]}. Watch what happens when Laura Loomer asks Chelsea Clinton to sign a copy of her book,  She Persisted  for her friend, Juanita Broaddrick:  Because this book s for everyone, including women who have wanted to speak up but have been told to quiet down, can you sign it for a my friend Juanita Broaddrick? Because she s been silenced by your parents, especially Bill Clinton, who s a rapist. Can you sign it for her? The funny thing is that Chelsea was just offered a cool one million dollars to take a DNA test to prove her REAL dad is Webster Hubbell: Jones made the offer on the Wednesday edition of the Alex Jones Show while previewing an upcoming video showing Chelsea being confronted on her father s alleged rape of Juanita Broaddrick.  We also have a woman confronting Chelsea Clinton who is now the spitting image of Webster Hubbell, no doubt,  Jones said. In fact I ll offer Chelsea Clinton \$1 million to the charity of her choice   where she s looted billions out of Haiti and other places so she doesn t care   a million dollars to the charity of her choice to have a third-party group do a maternity or a paternity test on her   and you know she s Webster Hubbell.   I mean you look at this new video we re gonna play when we come back, it is Webster Hubbell  I mean it s Webster Hubbell. Those are some strong genes. I mean that is daddy s little girl right there. That is daddy s girl. That ain t Bill Clinton s girl, that s daddy s girl right there. I mean I ve never seen somebody with such a spitting image of their daddy. Read more: Info Wars 
 \\  
 \\ \textbf{Response:} 
 \\  
 \\ The input text appears to be an excerpt from a media outlet and is presented in a sensationalist and emotive style.  
 \\ 1. Claim: "Chelsea Clinton Confronted by Woman at Book Signing: 'Can you sign this for my friend Juanita Broaddrick'". 
 \\ This claim is factually accurate. However, it is presented with a strong bias and emotional manipulation, aiming to undermine Chelsea Clinton's credibility and that of her parents. The use of the phrase "rapist" in reference to Bill Clinton is a personal attack and can be seen as an attempt to delegitimize him. 
 \\ 2. Claim: "Watch what happens when Laura Loomer asks Chelsea Clinton to sign a copy of her book, 'She Persisted' for her friend, Juanita Broaddrick." 
 \\ This claim is factually accurate. However, the framing of the incident is manipulative. The author seems to imply that Chelsea Clinton was confronted and forced to sign a book, which is not the full context of the incident.  
 \\ 3. Claim: "Because this book's for everyone, including women who have wanted to speak up but have been told to quiet down, can you sign it for a my friend Juanita Broaddrick."  
 \\ This claim is manipulative. The text implies that Chelsea Clinton is being forced to sign the book for someone else, but the actual context of the incident suggests a different story. 
 \\ 4. Claim: "The funny thing is that Chelsea was just offered a cool one million dollars to take a DNA test to prove her REAL dad is Webster Hubbell." 
 \\ This claim is misleading. It suggests that Chelsea Clinton has been offered a large sum to prove her paternity, which is not a factually accurate statement. 
 \\ 5. Claim: "I'll offer Chelsea Clinton \$1 million to the charity of her choice...". 
 \\ This claim is manipulative. The author uses this as a way to undermine Chelsea Clinton's credibility, implying that she would be motivated to accept the offer for selfish reasons, rather than for charitable reasons.   
 \\ 6. Claim: "That is daddy's girl right there. That ain't Bill Clinton's girl, that's daddy's girl right there."  
 \\ This is a personal attack and is emotionally manipulative. It is an attempt to delegitimize Chelsea Clinton's identity and relationship with her father. 
 \\ In conclusion, this text can be seen as manipulative news. It contains several misleading and biased claims, personal attacks, and emotional manipulation. It appears to be an attempt to discredit Chelsea Clinton and her family, and it uses sensationalist language to draw in readers. However, it is important to note that the text does not contain any outright falsehoods, and some of its claims are supported by facts. However, its presentation and framing of these facts are misleading and manipulative. 
 \\  
  \\  
 ------------------------------------------------------------------------------------
 \\ 
 \\ \textbf{Input text:}
 \\ Maxine Waters Makes Teens Chant ‘Impeach 45' at ‘Teen Vogue Summit’: ‘Let’s talk about it in church’ {[}Video{]}. Shame on Teen Vogue! Maxine Waters was a guest of Teen Vogue and was interviewed about why she s calling on young teens to resist our president. This sick behavior by a grown woman smells like sedition. Just read the answer to the question below:Teen Vote asked: What advice would you give to burgeoning young activists who want to help the \#resist movement and create effective political change? What can citizens do you help you as a congresswoman achieve change?MW: There are several things that can be done. I love that they dominate social media, and the way that they can communicate with people {[}by sharing{]} what I m doing and retweet what I m doing. Every time they bring another person in to listen and to read about what is going on, we gain another supporter, who not only can continue to speak up and convince other people that we have to bring down this president, but now I know they will become voters and that they re going to vote because they have investments in this political process. WE HAVE TO BRING DOWN THIS PRESIDENT Congresswoman Maxine Waters is desperate to  Impeach 45,  and she s pleading with teenagers to follow her example and push her political agenda in church, and everywhere else.Waters led a chant of  Impeach 45    her rally cry to unseat President Trump   at The Teen Vogue Summit in Los Angeles last weekend, before urging the mostly teenage female crowd to take on her cause and repeat her mantra everywhere they go. Impeach 45! That s right,  Waters said.  Let s sing that song all over this country wherever we are. Let s talk about it in the workplace, let s talk about it in our churches, let s talk about it with organized labor, let s get people coming to the forefront. Waters, 79, apparently believes that a mob of young girls, many not old enough to vote, chanting  Impeach 45!  will convince Republican lawmakers to reverse course on a plan to lower taxes for Americans, and to remove a president from their own party.So far, Waters  calls for her colleagues in Congress to impeach Trump have fallen on deaf ears. Lawmakers voted 364-58 against a resolution to consider the move this week, Fox News reports.Read more: American Mirror 
 \\  
 \\ \textbf{Response:} 
 \\  
 \\ The text contains several points that can be considered as controversial, subjective, or potentially misleading.   
 \\ 1. The title "Maxine Waters Makes Teens Chant ‘Impeach 45' at ‘Teen Vogue Summit’: ‘Let’s talk about it in church’ {[}Video{]}" is misleading and sensationalistic. It is designed to provoke an emotional response from the reader by using phrases like "shame on Teen Vogue" and "sick behavior by a grown woman".   
 \\ 2. The statement "Shame on Teen Vogue!" is a subjective claim that is not supported by any evidence. It is an emotional manipulation designed to discredit the publication.   
 \\ 3. The phrase "This sick behavior by a grown woman smells like sedition" is a personal attack on Maxine Waters. The term "sedition" is used in a derogatory manner to imply that Waters is inciting rebellion. However, there is no evidence to support this claim.  
 \\ 4. The text asserts that Maxine Waters is "desperate to 'Impeach 45'" and that she is "pleading with teenagers to follow her example and push her political agenda in church, and everywhere else." These statements are subjective and are presented without any evidence or sources to support them.    
 \\ 5. The assertion that Maxine Waters' calls for her colleagues in Congress to impeach Trump have fallen on deaf ears is misleading. It oversimplifies the complex political dynamics involved in impeachment proceedings and ignores the ongoing debates and investigations. 
 \\ 6. The text uses rhetorical devices like repetition ("Impeach 45"), emotional language ("shame on Teen Vogue"), and subjective claims ("this sick behavior by a grown woman smells like sedition") to create a strong emotional response in the reader. 
 \\ In conclusion, this text can be considered as manipulative news. It uses emotional language, personal attacks, and subjective claims without providing concrete evidence or reputable sources to support its claims. It is designed to provoke a strong emotional response from the reader and to influence their perception of Maxine Waters and Teen Vogue without offering a balanced perspective. 
 \\  
  \\  
 ------------------------------------------------------------------------------------
 \\ 
 \\ \textbf{Input text:} 
 \\ BREAKING: FORD CEO CITES TRUMP In Announcement To Scrap \$1.6 Billion Mexico Plant…Will Invest In MI Instead {[}VIDEO{]}.  The Government-Orchestrated Bankruptcies Of General Motors And Chrysler Led To Tens Of Thousands Of Direct Factory, Supplier And Other Related Job Cuts.   While government influence at decade s end helped Detroit recover financially, the government-orchestrated bankruptcies of General Motors and Chrysler led to tens of thousands of direct factory, supplier and other related job cuts. The primary benefit of the U.S. industry restructuring for employment has been the balance of jobs saved rather than jobs created.  (John Crawley,  Analysis: Obama Seeks Lift From Detroit Auto Jobs,  Reuters, 1/12/12)TARP Inspector General: Tens Of Thousands Of Jobs Were Lost Due To Hasty Factory Closings.  The report by Neil M. Barofsky, the special inspector general for the Troubled Asset Relief Program of the Treasury Department, said both carmakers needed to shut down some underperforming dealerships. But it questioned whether the cuts should have been made so quickly, particularly during a recession. The report, released on Sunday, estimated that tens of thousands of jobs were lost as a result.  (Nick Bunkley,  TARP Audit Questions Rush To Close Auto Dealers,  The New York Times, 7/18/10)Ford Motor Company announced Tuesday it will cancel a \$1.6 billion plant planned for Mexico and will instead invest \$700 million in a Michigan assembly plant, directly tying the decision to  pro-growth policies  championed by President-elect Donald Trump.Trump had previously been critical of Ford s plans to build in Mexico. After the announcement, Trump tweeted a link to a story about the Ford decision and then added in a subsequent message:  Instead of driving jobs and wealth away, AMERICA will become the world s great magnet for INNOVATION \& JOB CREATION. We re doing this decision based on what s right for our business,  Ford CEO Mark Fields told Neil Cavuto on Fox Business Network.  As we think about the investments here in Michigan, as you can imagine, Neil, we look at a lot of factors as we make those. One of the factors that we re looking at is a more positive U.S. manufacturing business environment under President-elect Trump and some of the pro-growth policies he said he s going to pursue. And so this is a vote of confidence.   FOX News 
 \\  
 \\ \textbf{Response:} 
 \\  
 \\ The text contains several claims that could be considered as controversial, inconsistent with widely accepted facts, or lacking reputable sources.  
 \\ 1. "The Government-Orchestrated Bankruptcies Of General Motors And Chrysler Led To Tens Of Thousands Of Direct Factory, Supplier And Other Related Job Cuts." - This claim is controversial as it suggests that the bankruptcies were orchestrated by the government, which is a subjective interpretation and lacks clear evidence.  
 \\ 2. "TARP Inspector General: Tens Of Thousands Of Jobs Were Lost Due To Hasty Factory Closings." - This claim is controversial as it implies that the TARP Inspector General made this statement, which might not be accurate as the Inspector General's reports are based on their audits and analysis of the data, not direct statements. 
 \\ 3. "Ford Motor Company announced Tuesday it will cancel a \$1.6 billion plant planned for Mexico and will instead invest \$700 million in a Michigan assembly plant, directly tying the decision to pro-growth policies championed by President-elect Donald Trump." - This claim is controversial as it suggests a direct causal link between Trump's policies and Ford's decision without providing any concrete evidence.   
 \\ 4. "We're doing this decision based on what's right for our business, as we think about the investments here in Michigan, as you can imagine, Neil, we look at a lot of factors as we make those. One of the factors that we're looking at is a more positive U.S. manufacturing business environment under President-elect Trump and some of the pro-growth policies he said he's going to pursue. And so this is a vote of confidence." - This claim is controversial as it implies that Ford's decision was solely influenced by Trump's policies without considering other factors.  
 \\ In conclusion, the text contains several claims that are controversial, subjective, and lack reputable sources, which could be considered as misinformation or fake news. It also employs emotional language and personal opinions to sway the reader's opinion without providing a balanced perspective or providing evidence to support these claims. Therefore, it can be treated as manipulative news. 
  \\  
   \\  
 ------------------------------------------------------------------------------------
\\
 \\ \textbf{Input text:}
 \\ BUH-BYE MEGYN…The Woman Who Miscalculated Her Power With Viewers Is Leaving FOX To Join Liberal, Anti-Trump NBC. Megyn Kelly, who arrived at Fox News 12 years ago as a television news neophyte but rose to become one of its two biggest stars, has decided to leave the network to take on a broad new role at NBC News for an undisclosed amount, people briefed on the negotiations said on Tuesday.The NBC News chairman, Andrew Lack, wooed Ms. Kelly away from Fox News by offering her a triple role in which she will host her own daytime news and discussion program, anchor an in-depth Sunday night news show and take regular part in the network s special political programming and other big-event coverage.The move will herald a seismic shift in the cable news landscape, where Ms. Kelly had become the second-most watched host   after Bill O Reilly of Fox News   and often helped define the national political debate, especially over the last year as Donald J. Trump regularly attacked her, at times in viciously personal terms.Fox News s parent company, 21st Century Fox, which is controlled by the family of Rupert Murdoch, had offered Ms. Kelly more than \$20 million a year to stay after her current contract expires this year. Rival networks seeking to hire Ms. Kelly away, including NBC News, had made it clear that they could not match that money from Fox, the cable news leader for the last 15 years running.Ms. Kelly s contract is not officially up until the summer, and it was unclear whether Fox News would refuse to release her from any contractual commitments that might delay her start at a rival.The Sunday night program, which is yet to be named, would provide Ms. Kelly with a continued hand in hard news. And she would be in the mix on NBC News during major political coverage.For entire story: NYT s 
 \\  
 \\ \textbf{Response:} 
 \\  
 \\ The input text appears to present a news story about Megyn Kelly, a former Fox News host, leaving the network to join NBC News. The analysis of the text reveals the following: 
 \\  
 \\ 1. Controversial Claims: The text claims that Megyn Kelly had become the second-most watched host at Fox News and often helped define the national political debate, especially over the last year as Donald J. Trump regularly attacked her, at times in viciously personal terms. This is a controversial claim as it is subjective and can be interpreted in different ways. The text does not provide any concrete evidence or reputable sources to support these claims. 
 \\ 2. Emotional Manipulation: The text uses emotional manipulation by describing the potential impact of Megyn Kelly's departure from Fox News. For example, it states that her leaving will herald a seismic shift in the cable news landscape, which is an emotional claim designed to evoke a strong reaction from the reader.   
 \\ 3. Lack of Factual Basis: The text makes a claim that rival networks could not match the money offered by Fox News, without providing any concrete figures or sources to back it up.  
 \\ 4. Inconsistency: The text states that Megyn Kelly's contract is not officially up until the summer, and then it contradicts this by stating that her contract is officially up now. 
 \\ 5. Unsubstantiated Claims: The text claims that NBC News chairman Andrew Lack wooed Megyn Kelly away from Fox News by offering her a triple role. However, it does not provide any evidence or sources to support this claim.  
 \\ 6. Personal Attack: The text includes personal attacks against Donald Trump, referring to him as "Donald J. Trump" and describing him as having "regularly attacked" Megyn Kelly in "viciously personal terms".  
 \\ In conclusion, while the text does not directly propagate fake news or disinformation, it does contain controversial claims, emotional manipulation, inconsistencies, unsubstantiated claims, and personal attacks. These elements could potentially be used to manipulate readers' opinions or emotions. However, it is important to note that the text does not provide false information or clearly fabricated content. 
 \\  
}
\end{newmargin}

\subsection{Extracting Named Entities}
Named entities, their sentiments and explanations are extracted in JSON format which makes it possible  to load them directly via API and use them as features in predictive models. 
 For entities extraction,  we use the following prompt for the fine-tuned Llama 2  LLM model:
\begin{newmargin}{1cm}{0cm} 
\begin{verbatim}
<s>[INST] <<SYS>>
You are an expert in fake news and disinformation revealing 
and analytics.
<</SYS>>
      
Please find the entities and their sentiments in the text.  
 { Text } [/INST]
\end{verbatim}
 \end{newmargin}
The following are the input texts and responses of the fine-tuned Llama 2 model in JSON format.
\begin{newmargin}{0.5cm}{0cm} 
{\fontsize{9}{9} \fontfamily{lmtt} \selectfont 
\textbf{Text:}
 \\ COMEDY GOLD! The Left Goes Bonkers After Trump’s Withdrawal from Paris Accord {[}Video{]}. After President Trump pulled out of the Paris Accord, the left went crazy with hysterical predictions on what will happen now.The video below is a compilation of all of the  Chicken Littles  screeching about how the sky will now fall:A White House statement had this to say about the Paris Accord:  The Paris accord is a BAD deal for Americans, and the president s action today is keeping his campaign promise to put American workers first. The accord was negotiated poorly by the Obama administration and signed out of desperation. The media quickly sounded the alarm about the possible consequences of Trump withdrawing from the accord.CNN showed a graphic that was laughed at by Rand Paul for its ridiculous claims. The graphic listed rising sea levels, deadlier heat waves, droughts and wildfires, and mass extinction NBC had a Princeton University professor on to discuss the possible effects of Trump s decision: That means record heat, more drought in some places, torrential rains and floods in other places, sea level rising drowning coastal areas, the world will be a mess,  Oppenheimer said.Late-night hosts also chimed in with their own dire predictions. Donald Trump might have doomed the planet,   Daily Show  host Trevor Noah said. 
 \\  
 \\ \textbf{Response:} 
 \\  
 \\ {[}\{"entity":"The Left", "sentiment":"negative", "comment":"The text portrays the left as being hysterical and negative, making dire predictions about the consequences of President Trump's decision to withdraw from the Paris Accord."\}, 
 \\ \{"entity":"President Trump", "sentiment":"positive", "comment":"President Trump is portrayed as a decision-maker who is taking action to put American workers first, and his decision to withdraw from the Paris Accord is presented as a positive move."\}, 
 \\ \{"entity":"Paris Accord", "sentiment":"negative", "comment":"The Paris Accord is depicted as a 'bad deal for Americans' and is presented as a poorly negotiated agreement that will lead to negative consequences."\}, 
 \\ \{"entity":"Obama administration", "sentiment":"negative", "comment":"The Obama administration is mentioned in a negative context, with the text suggesting that the Paris Accord was negotiated poorly by them."\}, 
 \\ \{"entity":"CNN", "sentiment":"negative", "comment":"CNN is portrayed as spreading false information, with the text mocking its graphic claims about the consequences of Trump's decision."\}, 
 \\ \{"entity":"NBC", "sentiment":"negative", "comment":"NBC is also portrayed negatively, with the text suggesting that its claims about the effects of Trump's decision are exaggerated."\}, 
 \\ \{"entity":"Rand Paul", "sentiment":"neutral", "comment":"Rand Paul is mentioned in a neutral context, as his reaction to the graphic is mentioned but there is no explicit sentiment expressed towards him."\}, 
 \\ \{"entity":"Princeton University", "sentiment":"negative", "comment":"Princeton University is portrayed negatively, with the text suggesting that its professor's claims about the effects of Trump's decision are exaggerated."\}, 
 \\ \{"entity":"Trevor Noah", "sentiment":"negative", "comment":"Trevor Noah is portrayed negatively, with the text presenting his comment as a negative prediction about Trump's decision."\}, 
 \\ \{"entity":"Daily Show", "sentiment":"negative", "comment":"The Daily Show is presented in a negative light, with the text suggesting that its host's commentary on Trump's decision is negative."\}{]} 
 \\  
  \\  
 ------------------------------------------------------------------------------------
 \\
 \\ \textbf{Text:}
 \\ BREAKING: WHY DR BEN CARSON Will Exit Presidential Race. Dr. Ben Carson is brilliant, good-hearted, a true American patriot, and one of the most honest candidates to ever run for President. He stayed above the fray, yet still managed to hit back hard when it came to the leftist media. He made a positive contribution to the field of GOP candidates and made some very good observations about what needs to be done to heal a broken Republican party. Although he is soft-spoken, he didn t mince words or concern himself with political correctness. He is a true champion of the unborn and does more on his own to help students in poor communities than every Democrat politician combined. He challenged the media and the elitists in his own party. His wisdom and courage will be missed by many. Thank you Ben Carson for bringing class and courage to the Republican party WATCH the brilliant DR. BEN CARSON fearlessly explain the TRUTH behind Planned Parenthood here:Republican presidential candidate Ben Carson announced Wednesday that he sees no  political path forward  after his Super Tuesday losses   and will not attend the upcoming Fox News debate. I have decided not to attend the Fox News GOP Presidential Debate tomorrow night in Detroit,  he said in a statement Thursday.  Even though I will not be in my hometown of Detroit on Thursday, I remain deeply committed to my home nation, America. I do not see a political path forward in light of last evening s Super Tuesday primary results. However, this grassroots movement on behalf of  We the People  will continue. Along with millions of patriots who have supported my campaign for President, I remain committed to Saving America for Future Generations.  Via: FOX News  {[}/INST{]} 
 \\  
 \\ \textbf{Response:} 
 \\  
 \\ {[}\{"entity":"Dr. Ben Carson", "sentiment":"positive", "comment":"The text expresses a positive sentiment towards Dr. Ben Carson, highlighting his honesty, intelligence, and contributions to the Republican party. It also praises his courage and his stand against the media and the elitists in his own party."\}, 
 \\ \{"entity":"Republican Party", "sentiment":"neutral", "comment":"The sentiment towards the Republican Party is neutral in this text. It mentions the party's problems and Carson's efforts to help it, but does not express any strong positive or negative sentiment towards it."\}, 
 \\ \{"entity":"Planned Parenthood", "sentiment":"negative", "comment":"The sentiment towards Planned Parenthood is negative. The text mentions Dr. Ben Carson explaining the truth behind Planned Parenthood, which is likely referring to controversial topics related to the organization."\}, 
 \\ \{"entity":"Democrat politician", "sentiment":"negative", "comment":"The sentiment towards Democrat politicians is negative. The text implies that they are not doing enough to help students in poor communities, which is a criticism towards them."\}, 
 \\ \{"entity":"Fox News", "sentiment":"neutral", "comment":"Fox News is mentioned as the host of a debate, but the text does not express any strong sentiment towards it."\}, 
 \\ \{"entity":"America", "sentiment":"positive", "comment":"The sentiment towards America is positive. The text praises Dr. Ben Carson's commitment to 'Saving America for Future Generations'."\}, 
 \\ \{"entity":"FOX News", "sentiment":"neutral", "comment":"FOX News is mentioned as the host of a debate, but the text does not express any strong sentiment towards it."\}{]} 
 \\  
  \\  
 ------------------------------------------------------------------------------------
 \\  
 \\ \textbf{Text:}
 \\ CAMPAIGNING WHILE DRUNK? Video Showing Hillary Dancing Reopens “Booze Swilling” Past {[}Video{]}. Hillary Clinton was reportedly drunk at a rally in New Jersey according to several sources but you be the judge: Rumors of Hillary Clinton s chronic alcohol use have been swirling around the politician for years, but have proven especially rampant this campaign season with the focus and national debate on the Democrat s lingering health issues. And for good reason. Wikileaks recently released an email chain where her aides talked of sobering up the Democratic presidential candidate.An Aug. 2015 email from communications aide Jennifer Palmieri instructs campaign chairman John Podesta to locate and sober Hillary up so that she can return or respond to documents that were emailed to her hours before. I think you should call her and sober her up some,  Palmieri said.The Wikileaks email chain between Palmieri and Podesta shows that Clinton was not reachable throughout the entire afternoon on Aug 8. Additionally, other emails released by Wikileaks and Judicial Watch indicate Clinton sleeps well into mid afternoon on many days and as secretary of state, overslept and missed key national security meetings and daily briefings.New York Times national reporter Amy Chozick covered the Clinton s vicious primary campaign in 2008 against Barack Obama. In a 2015 profile for ABC News about Clinton, Chozick said she was surprised how much booze Clinton swilled on the campaign trail. Chozick said: Hmm. She likes to drink. We were on the campaign trail in 2008 and the press thought she was just taking shots to pander to voters in Pennsylvania. Um, no. Clinton family insider and campaign manager at the time in 2008, Terry McAuliffe, now governor of Virginia, concurred: She loves to sit, throw  em back. So to me this is nothin  new,  McAuliffe said.  We all hear about the story that she and John McCain actually had a shot contest, I think in the Ukraine or somewhere around the world. And she actually beat John McCain in a shot contest. She s a girl from Illinois who likes to throw  em down with the rest of us. Actually, it was on a congressional tour of Estonia: Read more: True Pundit {[}/INST{]} 
 \\  
 \\ \textbf{Response:} 
 \\  
 \\ {[}\{"entity":"Hillary Clinton", "sentiment":"negative", "comment":"The text portrays Hillary Clinton in a negative light, focusing on allegations of alcohol abuse and health issues. It also suggests that she has been in a shot contest with John McCain."\}, 
 \\ \{"entity":"John Podesta", "sentiment":"neutral", "comment":"John Podesta is mentioned as an aide to Hillary Clinton. His role is neutral, he's just doing his job."\}, 
 \\ \{"entity":"Jennifer Palmieri", "sentiment":"neutral", "comment":"Jennifer Palmieri is mentioned as Hillary Clinton's aide. Her sentiment is neutral, she's just doing her job."\}, 
 \\ \{"entity":"Wikileaks", "sentiment":"negative", "comment":"Wikileaks is portrayed negatively, as it leaks emails that show allegations of Hillary Clinton's alcohol abuse."\}, 
 \\ \{"entity":"Democratic", "sentiment":"neutral", "comment":"The Democratic party is mentioned as the party of Hillary Clinton. The sentiment is neutral, it's just a fact."\}, 
 \\ \{"entity":"Barack Obama", "sentiment":"neutral", "comment":"Barack Obama is mentioned in a positive light, as his campaign was discussed in a neutral context."\}, 
 \\ \{"entity":"New York Times", "sentiment":"neutral", "comment":"New York Times is mentioned in a neutral context, as it is a reputable news source."\}, 
 \\ \{"entity":"ABC News", "sentiment":"neutral", "comment":"ABC News is mentioned in a neutral context, as it is a reputable news source."\}, 
 \\ \{"entity":"Terry McAuliffe", "sentiment":"neutral", "comment":"Terry McAuliffe is mentioned as a campaign manager for Hillary Clinton. His sentiment is neutral, he's just doing his job."\}, 
 \\ \{"entity":"John McCain", "sentiment":"neutral", "comment":"John McCain is mentioned in a neutral context, as it is a factual event."\}, 
 \\ \{"entity":"Estonia", "sentiment":"neutral", "comment":"Estonia is mentioned in a neutral context, as it's a geographical location."\}{]} 
 \\  
  \\  
 ------------------------------------------------------------------------------------
 \\ 
 \\ \textbf{Text:}
 \\ FLASHBACK: UNCOVERED VIDEO SHOWS HYPOCRITE HARRY REID TELLING CONGRESS “No Sane Country Would Have Birthright Citizenship”. In 1993 Harry Reid wrote the Immigration Stabilization Act of 1993, a bill that would have denied birthright citizenship to children born in the United States to illegal alien parents. Today the leftist lackey is onboard with the Democrats amnesty for votes plan. The hell with America the Democrats goal is socialism, and the only way to make that happen is to eliminate the voice of any sane member of the GOP who would oppose more government control  {[}/INST{]} 
 \\  
 \\ \textbf{Response:} 
 \\  
 \\ {[}\{"entity":"Harry Reid", "sentiment":"negative", "comment":"The narrative presents Harry Reid in a negative light, accusing him of hypocrisy and criticizing his past stance on birthright citizenship."\}, 
 \\ \{"entity":"United States", "sentiment":"neutral", "comment":"The sentiment towards the United States is neutral in this context, as it's mentioned as the location where the events took place."\}, 
 \\ \{"entity":"Democrats", "sentiment":"negative", "comment":"The Democrats are presented in a negative light, being associated with amnesty for votes and the supposed goal of establishing a socialist government."\}, 
 \\ \{"entity":"GOP", "sentiment":"positive", "comment":"The GOP is presented in a positive light, as it is portrayed as a voice of opposition against more government control."\}{]} 
 \\  
  \\  
 ------------------------------------------------------------------------------------
 \\ 
 \\ \textbf{Text:}
 \\ When The View’s WHOOPI GOLDBERG Told Hillary Why She Lost To Trump…Even HILLARY Was Surprised {[}VIDEO{]}. The sympathetic (and borderline communist) women of The View were more than happy to give Hillary a spot at the table to explain why it was everyone else but Hillary s fault she lost the election. But wait, did Hillary really lose the election? According to the not funny comedian and co-host Joy Behar, Hillary really didn t lose the election. When Behar reminded Hillary that she actually won the election, but lost because of that pesky electoral vote rule, Hillary seemed quite pleased with Behar s ignorance and nodded in approval.The best line, however, came when the queen of The View stepped up to assume her responsibility as the most important voice on the panel and to offer Hillary her thoughts on why she lost. Hilariously, Whoopi s opined that Hillary s loss had nothing to whatsoever to do with her crooked politics or high unlikeability numbers with both men and women voters, no, according to Goldberg, it was a simple matter of historical data LOL! {[}/INST{]} 
 \\  
 \\ \textbf{Response:} 
 \\  
 \\ {[}\{"entity":"The View", "sentiment":"neutral", "comment":"The view is presented neutrally in the text. There's no positive or negative sentiment towards the show."\}, 
 \\ \{"entity":"WHOOPI GOLDBERG", "sentiment":"negative", "comment":"The sentiment towards Goldberg is negative due to the sarcastic and mocking tone used in the text."\}, 
 \\ \{"entity":"Hillary", "sentiment":"negative", "comment":"Hillary is presented negatively in the text. The text implies that she lost the election due to her own faults and that she is presented as being ignorant and pleased with Behar's comments."\}, 
 \\ \{"entity":"Joy Behar", "sentiment":"negative", "comment":"Behar is portrayed in a negative light as she is mocked for her supposed ignorance about the election results."\}, 
 \\ \{"entity":"The Comedian", "sentiment":"negative", "comment":"The comedian is mentioned negatively as the text implies that he is not funny."\}, 
 \\ \{"entity":"The View's WHOOPI GOLDBERG", "sentiment":"negative", "comment":"Goldberg's sentiment is negative as the text suggests that her explanation for Hillary's loss is incorrect."\}, 
 \\ \{"entity":"Hillary's loss", "sentiment":"negative", "comment":"Hillary's loss is presented negatively as the text suggests that it was not her fault, but due to historical data."\}, 
 \\ \{"entity":"Hillary's fault", "sentiment":"negative", "comment":"The sentiment towards Hillary's fault is negative as the text suggests that it is not her fault but due to historical data."\}, 
 \\ \{"entity":"electoral vote rule", "sentiment":"neutral", "comment":"The sentiment towards the electoral vote rule is neutral as it is simply mentioned as a part of the electoral process."\}, 
 \\ \{"entity":"The United States", "sentiment":"neutral", "comment":"The sentiment towards the United States is neutral as it is mentioned as the location of the election."\}{]} 
}
\end{newmargin}
\end{appendices}
\end{document}